%% file: root.tex
\documentclass[letterpaper, 10 pt, conference]{ieeeconf}  

\IEEEoverridecommandlockouts                              

\usepackage{epsfig} 
\usepackage{cleveref}
\crefname{figure}{fig.}{fig.}
\crefname{equation}{equ.}{equ.}

\usepackage{url}
\usepackage{multirow}
\usepackage{cite}
\usepackage{threeparttable}
\usepackage{algorithm}
\usepackage{algpseudocode} 
\usepackage{caption}
\usepackage{subcaption}
\usepackage{siunitx}
\usepackage{booktabs} 

\title{\LARGE \bf
Multi-UAV Disaster Environment Coverage Planning with Limited-Endurance}

 \author{
\thanks{$^ *$ These authors contribute equally.}
 Hongyu Song$^{* \dag}$,  Jincheng Yu$^{* \S}$,
Jiantao Qiu$^{\S}$, 
Zhixiao Sun$^{\P}$, Kuijun Lang$^{\P}$, Qing Luo$^{\P}$,
Yuan Shen$^{\S}$ \\and Yu Wang$^{\S}$
\thanks{$^ \dag$ Department
of Electrical and Computer Engineering, Technische Universität München, Munich, Germany. This work is during his intership at Tsinghua University.}
\thanks{$^ \S$ EE department, Tsinghua University. Beijing, China. \{yu-wang, yu-jc\}@tsinghua.edu.cn}
\thanks{$^ \P$ Avic Shenyang Aircraft Design And Research Institute. Shenyang, China.}
\thanks{This work is supported by Tsinghua EE Xilinx AI Research Fund and Tsinghua-Meituan Joint Research Institute.}
}

\begin{document}

\maketitle
\thispagestyle{empty}
\pagestyle{empty}

\begin{abstract}
\input{src/abstr1}
\end{abstract}

\section{Introduction}
\input{src/intro1}

\section{Related Work}
\input{src/relatedwork.tex}

\section{PROBLEM DEFINITION \& PRELIMINARIES}
\subsection{MAEl-CPP Problem}
\input{src/PROBLEMDEFINITIONPRELIMINARIES}
\section{Potential Disaster Heatmap}
\input{src/PotentialDisasterHeatmap.tex}

\section{Our Algorithms}
\input{src/OurAlgorithms.tex}

\section{RESULTS \& ANALYSIS}
\input{src/RESULTS}

\section{Conclusion \& Future Work}
\input{src/conclusion.tex}

\input{mael-cpp.bbl}
\bibliographystyle{IEEEtran}
\bibliography{mael-cpp}

\end{document}

%% file: src/abstr1.tex
Disaster areas involving floods and earthquakes are commonly large, with the rescue time being quite tight, suggesting multi-Unmanned Aerial Vehicles (UAV) exploration rather than employing a single UAV. For such scenarios, current UAV exploration is modeled as a Coverage Path Planning (CPP) problem to achieve full area coverage in the presence of obstacles. However, the UAV's endurance capability is limited, and the rescue time is constrained, prohibiting even multiple UAVs from completing disaster area coverage on time. Therefore, this paper defines a multi-Agent Endurance-limited CPP (MAEl-CPP) problem that is based on an \emph{a priori} known heatmap of the disaster area, which affords to explore the most valuable areas under UAV limited energy constraints. Furthermore, we propose a path planning algorithm for the MAEl-CPP problem by ranking the possible disaster areas according to their importance through satellite or remote sensing aerial images and completing path planning according to this ranking. Experimental results demonstrate that the search efficiency of the proposed algorithm is 4.2 times that of the existing algorithm.

%% file: src/intro1.tex
Multi-UAV exploration is more efficient than single UAV \cite{8485481} in disaster relief work where the area, such as in floods and earthquakes, to be explored is usually large and the rescue is time-limited \cite{9525446}.
Using the rescued survivor number and the time used as the evaluation metric, Yang et al.\cite{LSAR} proves that the average time to rescue each survivor significantly reduces as the number of UAVs increases.

\par Existing UAV exploration methods model the multi-UAV exploration task as a coverage path planning (CPP) problem \cite{inproceedings}, where the UAVs' starting point is arbitrary within a region containing no-fly zones and obstacles. However, as the UAVs cruise at least 30-40 meters from the ground \cite{balotuaadvanced}, a real-world rescue scenario almost without obstacles at that altitude force this type of CPP modeling to be inapplicable.

Additionally, a UAV's endurance is limited \cite{en11092221}, while the rescue time is quite constrained. Thus, despite utilizing multiple UAVs, complete on-time disaster area coverage is not guaranteed. For example, Zhengzhou flooded this year, with its Zhongyuan District being 193 square kilometers, requiring 20 UAVs operating for about 6.9 hours to achieve uninterrupted coverage. This is far more than the 1-hour no-load endurance of standard multi-rotor UAVs. At the same time, the 6.9 hours task time exceeds the safe survivor's survival time of specific disasters. Therefore, considering a UAV's limited endurance, it must explore places first where dangers are more likely to occur, e.g., the probability of a fire in a stadium is significantly greater than that of a lake. Therefore, we define a multi-Agent Endurance-limited CPP (MAEl-CPP) problem based on an \emph{a priori} known heatmap \cite{10.1007/978-3-030-58452-8_23} of the disaster area that affords to explore the most valuable areas (goals) under a limited energy strategy. Hence, trapped people or disaster-stricken property have a greater chance of being rescued and protected. Currently, pre-disaster maps/satellite images are exploited to detect and divide valuable areas. Nevertheless, calculating the optimal solution covering as many important goals as possible within a specific range driven by the UAV's endurance constrain is quite complicated \cite{8356712}. Using multiple robots to cover the target area is even more challenging. Simple path coverage algorithms, e.g., Zig-Zag continuous search \cite{10.1007/978-3-030-57802-2_64} do not distinguish targets, and multiple robots are prone to path repetition. Accordingly, the naive greedy algorithm will fall into a local optimum solution, and the robots will interfere with each other during the multi-robot collaboration prohibiting this method from solving the MAEl-CPP problem.

\begin{figure}[t]
    \centering
    \includegraphics[width=1\linewidth]{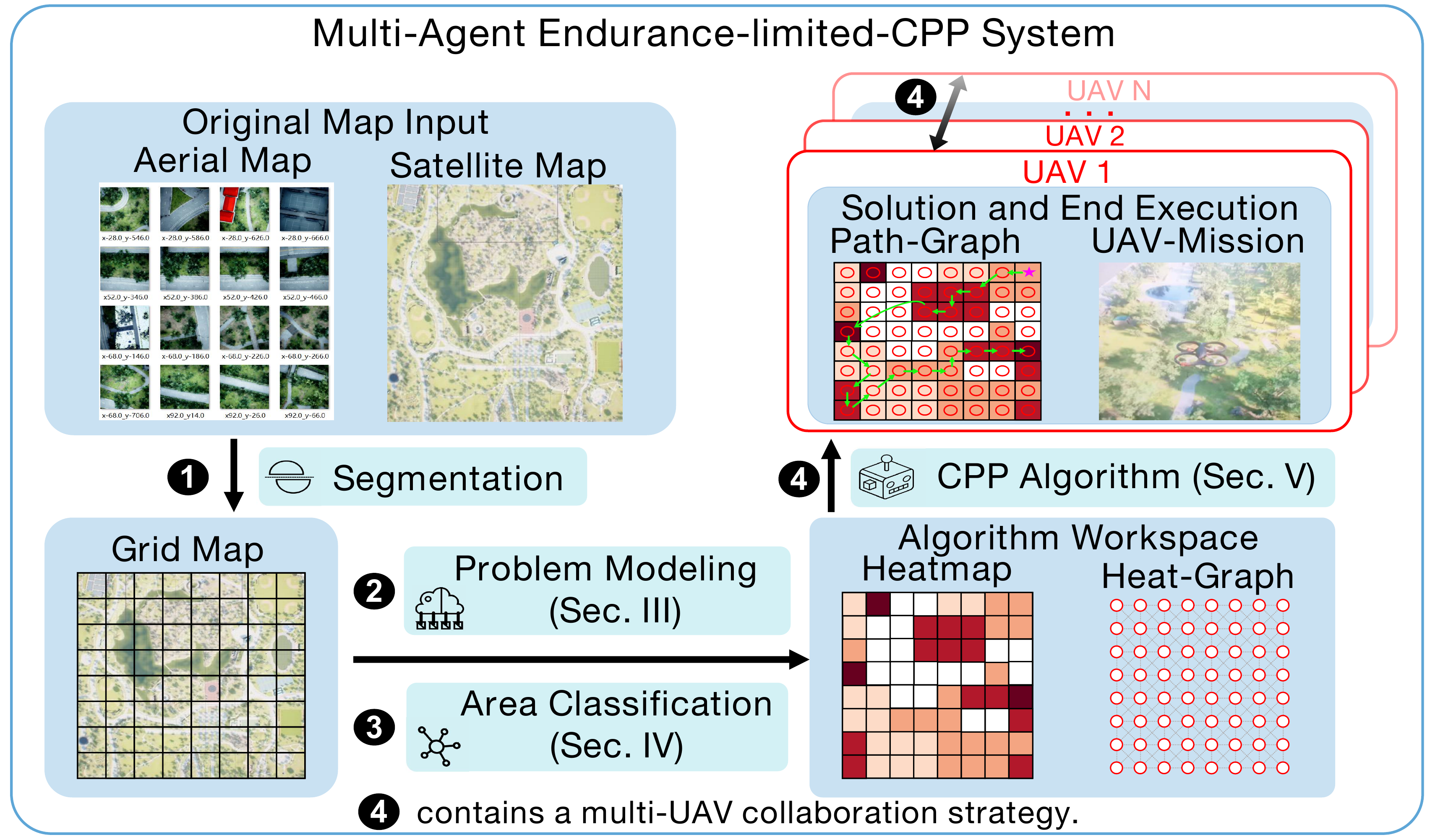}
    \caption{Overview of our MAEl-CPP framework.
    }
    \label{fig:sysframe}\vspace{-7mm}
\end{figure}
\par To address the above problems, we propose a multi-robot exploration framework illustrated in \Cref{fig:sysframe}. The contributions of our work can be summarized to:
\begin{itemize}
    \item Formulating a multi-Agent Endurance-limited CPP (MAEl-CPP) problem suitable for various potentially disaster-affected environments.
    \item A method generating \emph{a priori} heatmap of the pre-disaster environment exploiting a single environmental satellite map or a remote aerial map.
    \item We propose several path planning algorithms that are appropriate for MAEl-CPP. Among them, the SVReC presents an outstanding performance in search efficiency, reaching 4.2 times of Zig-Zag search.
\end{itemize}

\par The remainder of this paper is as follows. \Cref{sec:PROBLEM FORMULATION} presents our defined problem. Then, \Cref{sec:Potential Disaster heatmap} presents the method to convert the original map into a heatmap, while \Cref{sec:OUR ALOGORITHMS} generates the optimal path according to the \emph{a priori} heatmap. Then, based on the above problems and algorithms \Cref{sec:RESULTS&ANALYS} introduces the experimental setup. Finally, \Cref{sec:CONCLUSION} concludes this work and suggests future work.

%% file: src/relatedwork.tex
\label{sec:relatedwork}

Coverage path planning (CPP) has been widely discussed in the robotics community, the Backtracking Spiral Algorithm (BSA) \cite{BSA} is a classic work. This algorithm can completely cover an environment considering the occupied cells. Accordingly, Gabriely et al.\cite{gabriely2002spiral} propose a non-repetitive closed-loop coverage involving a closed environment containing obstacles based on competitive tours \cite{kalyanasundaram1994constructing}. M-STC*  \cite{tang2021mstc} expands the work of Gabriely et al. and suggests a solution that reasonably allocates the load distribution \cite{DBLP:journals/corr/abs-2107-14580} of multiple robots.  Torres et al.\cite{3D} develop a Zig-Zag-like path to complete the path coverage by targeting 3D reconstruction, which can work in convex and non-convex environments. Given that modeling these traditional CPP problems consider covering the entire map, these are not suitable for the MAEl-CPP problem. This is because MAEl-CPP aims to maximize the search efficiency rather than cover the entire map, as physical constraints make it challenging to cover it. 
 

Commonly, the map employed for the CPP problem follows a particular  distribution, e.g., BSA deals with path planning by dividing the grid cells into two types, covering and non-covering, or Panagou et al.\cite{panagou2016distributed} distinguish viewpoints into interested and uninterested. An et al. \cite{an2021distributed} introduce the prior distribution map to cover the area of interest. Accordingly, Lin et al.\cite{lin2009uav} utilize a probability map based on Gaussian distributions and suggest an algorithm solving the single-robot CPP problem for such a type of map. However, when the disaster occurs, the above maps cannot describe a particular area's specific \emph{a priori} conditions. The reason being the range of the real-world disaster hazard does not deliberately follow a specific distribution, and for the current types of maps, it is not trivial to deploy specific search efficiency-oriented tasks. Therefore, a map constructed according to the potential disaster risks of various real environments is mandatory for the MAEl-CPP problem.
 
Furthermore, some CPP algorithms adopt physical constraints. For example, Yang et al. limit the time searching for survivors, while Lin et al. use the time step of CPP to limit energy consumption, with each waypoint adding a time step. Similarly, m-STC* introduces $c$ as the robot's capacity, which is the sum of the total nodes in a spanning tree, with Franco et al.\cite{physical} considering a robot's more peculiar features. In any case, the UAV's energy consumption limit must be rigorously stated, but each UAV type has its efficiency and performance governing parameters. Hence, the energy consumption of the MAEl-CPP must be easily extended to different UAVs. Therefore, we define the endurance as the total path length calculated through the Euclidean distance between waypoints.

%% file: src/PROBLEMDEFINITIONPRELIMINARIES.tex
\label{sec:PROBLEM FORMULATION}
This section defines a new MAEl-CPP problem that sets the potential disaster level by utilizing representative features from the entire map. The disaster level should describe the probability of a specific disaster to occur in an area and conform to the common sense of search and rescue problems. If these requirements are fulfilled, we can deploy UAVs in a more realistic simulation or real environment. Then, according to the map's representative features, the grid map of the task space will be converted into a heatmap. The latter is a concise and informative map showing the differences between cells. Further details on deriving a heatmap from a grid map are introduced in \Cref{sec:Potential Disaster heatmap}.

\begin{figure}[tbp]
\centerline{\includegraphics[width=1\linewidth]{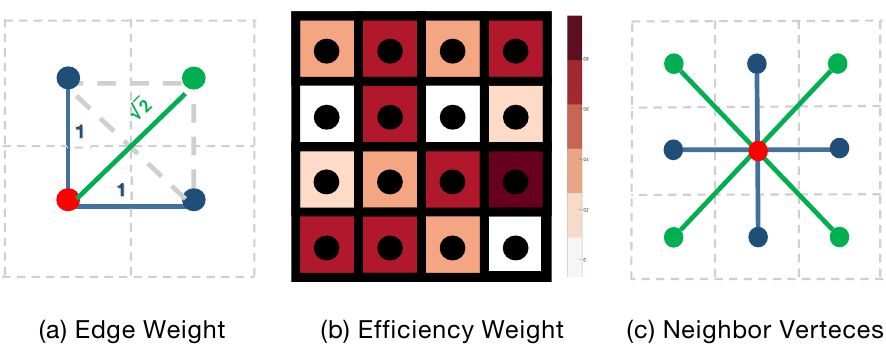}}
\caption{Edge Weight and Efficiency Weight. (a) Edge weight $e$ . The $e$ of the vertices adjacent to the side of the cell is 1, and the $e$ corresponding to the vertex whose cell position is diagonal is $\sqrt{2}$. (b) The relationship between the  Efficiency weight $r$ and the heat-value (color) of the cell. The higher the heat-value in the heatmap, the higher the Efficiency weight $r$ in the Heat-Graph. Further details on the heatmap are presented in Sec. IV. (c) The eight neighbor vertices of a vertex, where red is the current vertex, green and blue are the neighbor vertices of the diagonal and translation, respectively. }
\label{fig:weights}
\end{figure}

In the problem examined here, we employ a graph structure \cite{8968151} called Heat-Graph $\mathcal{H}$, with its vertices corresponding to the heatmap cells on a one-to-one basis that share the same positional relationship (\Cref{fig:weights}). For example, given cell $C_{xy}$ and vertex $\eta_{ij}$, whose vertex weight is called Efficiency weight $r$, xy denote the coordinates of cell center and ij is the row and column in the graph structure, ${r(\eta_{ij})}$ is mapped from the heat-value ${h(C_{xy})}$ of the heatmap illustrated in \Cref{fig:weights} (b). As a physical limitation, we stipulate that the endurance (Maximum path length) of each UAV is $\mathcal{D}$. The edge weight $e(\eta_{ij},\eta_{kl})$ between neighbor vertices is presented in \Cref{fig:weights} (a). For a heatmap with grid size of 40mx40m, we obtain:
\begin{equation}
    \eta_{ij} = \mathcal{G}(C_{xy}), i = (y_{max}-y)/40 + 1, j = x/40 + 1
\end{equation}
\begin{equation}
    e(\eta_{ij},\eta_{kl}) = \sqrt{(i-k)^2 + (j-l)^2}
\end{equation}
\begin{equation}
    r(\eta_{ij}) =\left\{
\begin{array}{rcl}
10      &     & {h(C_{xy}) = 1.0}\\
10h(C_{xy}) -2    &     & {else}
\end{array} \right.
\label{eq:Eff-weight}
\end{equation}
where $\mathcal{G}$ is the map created after converting the heatmap to the Heat-Graph, and the max accumulated edge weight is  $d_m=D/40$. We map ${h(C_{xy})}$ to ${r(\eta_{ij})}$ utilizing \Cref{eq:Eff-weight} to enhance the importance of one class and impose one of the remaining classes not to be covered.

Given $\mathcal{H}$ and $n$  UAVs, the flight trajectory of the $i$th UAV $U_i$ is $\Gamma_i=\mathcal{G}(\gamma_i)$ , where $\gamma_i$ is a directed sub-graph of $\mathcal{H}$ and contains $M$  $\eta$s. $R_i$ and $d_i$ are the accumulated Efficiency weight (Eff-weight) and edge weight, respectively. $R$ can be used to describe the overall search efficiency of a path.  Then  solving the path-graph $\gamma_i$ process can be summarized as:
\begin{equation}
\mathop{\arg\max}\limits_{\gamma_i} R_i, \quad \mathrm{ s.t.} i \leq n, d_i \leq d_m \\
\end{equation}
\begin{equation}
  R_i = \sum_{m=1}^Mr(\eta_m), d_i = \sum_{m=1}^{M-1}e(\eta_m,\eta_{m+1}), \quad \eta_m \in \gamma_i \\
\end{equation}

\subsection{Naive Greedy Path Coverage For a Single UAV}
After briefly describing the path coverage problem, we first introduce a relatively simple solution inspired by the greedy algorithm \cite{Greedy}, which is a classic method to solve NP-Hard problems \cite{8962157}. Specifically, we suggest the Naive Greedy (Na-Greedy) Path Coverage algorithm involving the following calculation steps. Given a Heat-Graph $\mathcal {H}$, a UAV start vertex $\eta_s$ and the max accumulated edge weight ${d_m}$, we obtain after path planning $\gamma_i$, $R$, and the cover rate $\sigma$, which is the percentage of covered vertices out of all vertices. The pseudo-code of the Na-Greedy algorithm is presented in \Cref{fig:Algorithm Overview} Algorithm 1.


\par Although the proposed Na-Greedy algorithm can quickly solve some MAEl-CPP problems from a Heat-Graph, in some cases, due to the hard constraints of the CLOSE set \cite{luan2021c}, which is the set of covered vertices. The UAV path may fall into a local optimal solution, i.e., all vertices around the UAV are in the CLOSE, and the UAV can not continue to plan the path after exiting the loop. To solve this problem, we propose the Heu-Greedy and SVReC algorithms introduced in \Cref{sec:OUR ALOGORITHMS}.

%% file: src/PotentialDisasterHeatmap.tex
\label{sec:Potential Disaster heatmap}
We efficiently generate the proposed map and assign different Eff-weights (\Cref{sec:PROBLEM FORMULATION}) to the cells containing different scenes, e.g., water and house scenes. Utilizing the proposed map for the actual MAEl-CPP problem requires building the map such as each cell is based on the probability of the occurrence of the actual disaster. However, it is challenging to create such a map, as most current maps considering the CPP problem focus on the grid occupancy \cite{o2018variable} or the probability distribution as described in \Cref{sec:relatedwork}. Thus, an image map from the real environment cannot be converted into the desired map type, and the existing map types are inefficient in searching and rescuing the post-disaster environment. The latter is due to the lack of current map types' of comprehending the potential disaster probability of various scenarios and the corresponding danger level after the disaster.

\begin{figure}[t]
\centering
\includegraphics[width=\linewidth]{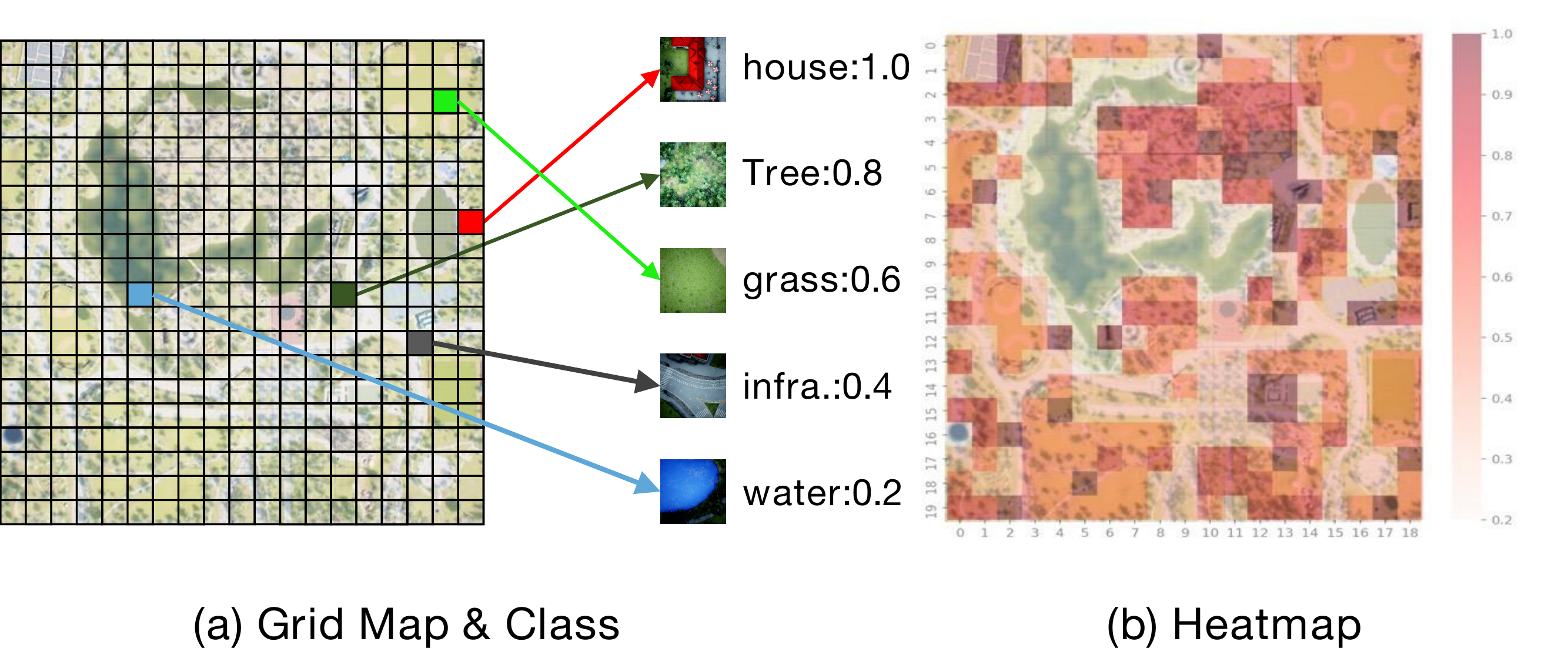}
\caption{Grid-cell Classification \& Heatmap. Classification is described next to the sample picture in the form of \textbf{class: heat-value}.}
\label{fig:classification}
\end{figure}

Additionally, the map must also guide the UAV's low-altitude flight during the search and rescue process. Therefore, we segment into a grid map a high-altitude satellite image \cref{fig:classification} (a) of our simulation scene. The grid map cells are classified into five classes according to the possibility that people in the scene are trapped or in danger when the fire occurs. It should be noted that classification is based on common sense, while several other criteria can be utilized depending on the scenario or the type of disaster, affording a universal classification concept.
 
Once the heatmap described above is generated, the first challenge is classifying the cells. Large-area city maps often contain tens of thousands of cells that need to be classified. Considering the difficulty of manual labeling, we manually add scene tags to only 20\% of the total grid map cells. Then we employ the pre-trained DenseNet121 \cite{huang2018densely} for feature extraction and Support Vector Machines (SVM \cite{li2004multilabel}) for classification. OpenCV \cite{opencv_2021} is also adopted to improve classification accuracy. After classification, the class labels are converted into heat values, and the grid map is converted into a heatmap affording concisely and efficiently distinguishing the importance of different cells. \Cref{fig:classification} (a) illustrates examples of the five classes and their corresponding heat-value in a heatmap, while \Cref{fig:classification} (b) presents our heatmap.


%% file: src/OurAlgorithms.tex
\begin{figure*}[htp]
\centerline{\includegraphics[width = \textwidth]{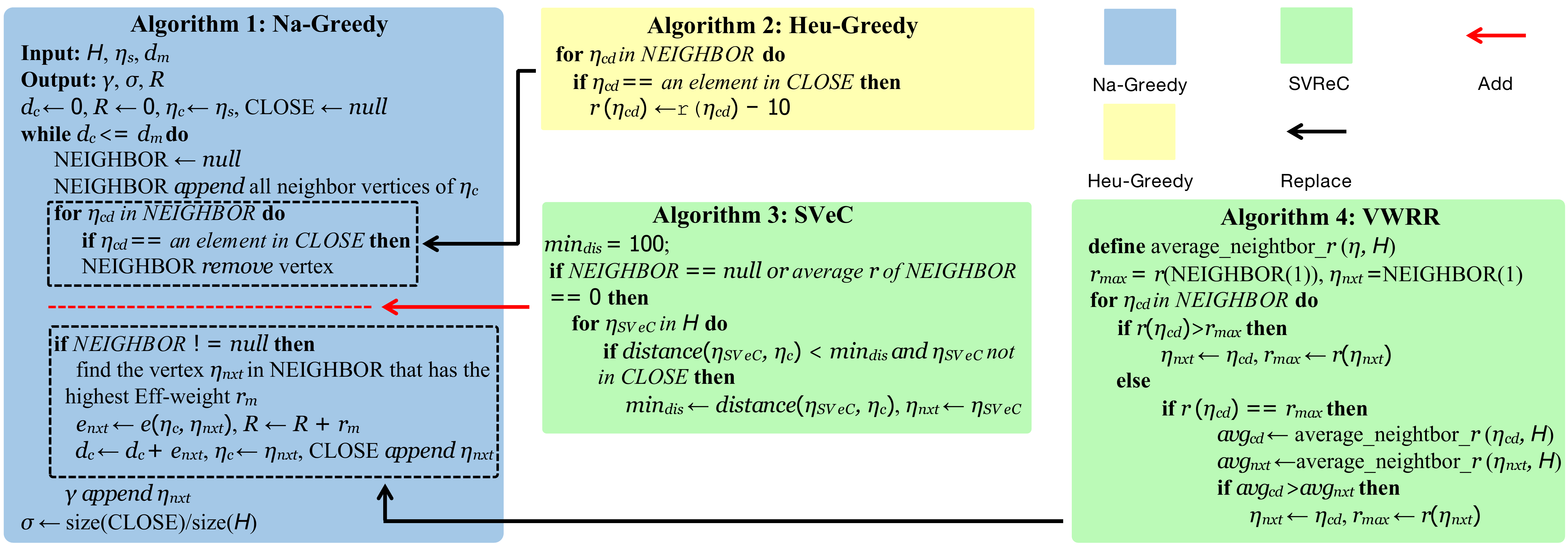}}
\caption{Algorithm Overview}
\label{fig:Algorithm Overview}
\end{figure*}

\label{sec:OUR ALOGORITHMS}
The Naive Greedy (Na-Greedy) algorithm is a processing method based on the vertex’s Eff-weight. It suffers from a Weight Trap problem when employed on a single-UAV CPP scheme and a Weight Interference problem when utilized for multi-UAV CPP. Both problems force ending the coverage planning loop because the UAVs cannot find the next target vertex. The simplest solution is adding heuristic weights and developing the Heu-Greedy algorithm presented in \Cref{fig:Algorithm Overview} Algorithm 2. The corresponding 
idea is as follows, whenever a candidate for $\eta_{nxt}$ is in CLOSE, a penalty of -10 is applied to its $r$ to assist in covering a known vertex but without accumulating weight, preventing it from exiting the loop when a known vertex cannot be covered. Additionally, this strategy affords to distinguish each covered vertex. However, it reduces the search efficiency due to the repeated coverage, which is termed dynamic Weight Traps (repeatedly beating between the covered vertices). Other notations are listed below: current vertex $\eta_c$, candidate vertex $\eta_{cd}$, potential target vertex $\eta_{SVeC}$, acumulated Eff-weight $R$, distance of the nearest sufficient vertex $min_{dis}$.

\begin{figure}[t]
\centerline{\includegraphics[width = \linewidth]{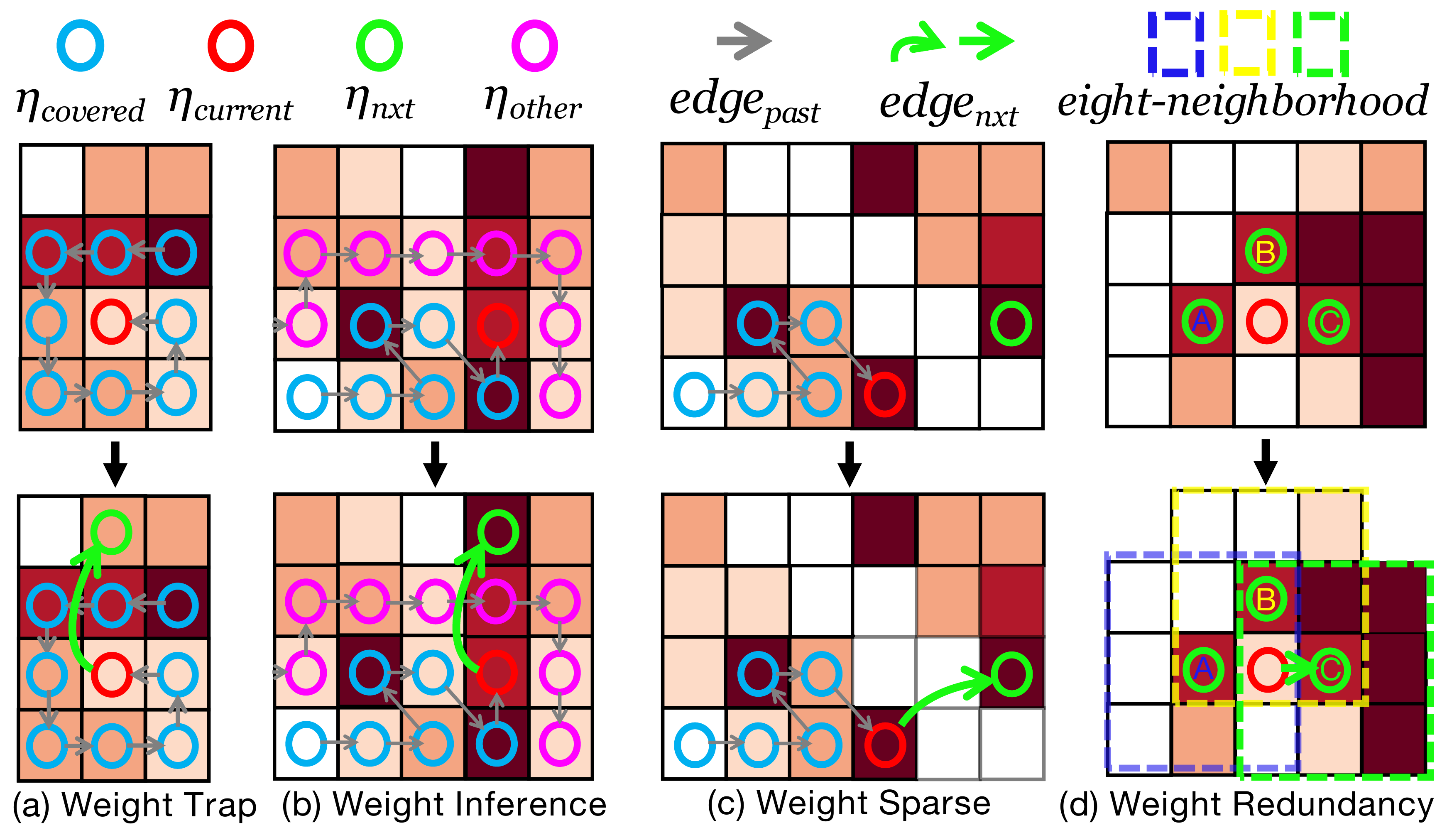}}
\caption{Problems Overview}
\label{fig:Problems Overview}
\end{figure}

Additionally, both Na-Greedy and Heu-Greedy algorithms suffer from the problems of Weight Sparse and Weight Interference, which will be introduced later. Although these problems do not affect coverage, they inject some invalid coverage. Hence, SVReC can be extended to a multi-UAV version by solving Weight Interference. We adopt the skip vertices coverage (SVeC) and the vertex weight resolution reinforcement (VwRR) methods to solve the above problems. Therefore, the overall algorithm is called SVReC. An overview of the above problems is illustrated in \Cref{fig:Problems Overview}.

\subsection{Weight Trap}

Weight Trap occurs when all reachable eight-neighborhood vertices of the current vertex have been covered. The reachable eight-neighborhood vertices are the neighbor vertices that do not exceed the map range. The Weight Trap also takes place when we use heuristic weights as soft constraints. A UAV can choose the vertices that have already been covered as potential paths because as the $r$ of vertices alternately descends, the vertices alternately become the next vertex in the path. As a result, the UAV ineffectively covers the area between these vertices. An example of a Weight Trap is depicted in \Cref{fig:Problems Overview} (a) above.
\par  The Weight Trap problem can be solved utilizing SVeC, with the corresponding pseudo-code shown in \Cref{fig:Algorithm Overview} Algorithm 3. When the reachable eight-neighborhood vertices of the current vertex are all covered, the next UAV target $\eta_{nxt}$ is specified as the nearest vertex to the current vertex, whose $r$ exceeds a certain threshold $r_{th} $. Solving the Weight Trap problem is presented in \Cref{fig:Problems Overview} (a) below.


\subsection{Weight Interference}

Similar to Weight Trap, Weight Interference occurs when several UAVs collaborate to cover adjacent vertices. These vertices surround the current vertex of a UAV during multi-UAV coverage. In turn, during this case, the coverage stops or loops between the vertices that have been covered in (\Cref{fig:Problems Overview} (b) above).

SVeC can solve this problem as a collaborate strategy. Nevertheless, it is different when multiple UAVs are used, as all UAVs share the same CLOSE set during planning. The corresponding solution is illustrated in \Cref{fig:Problems Overview} (b) below.

\subsection{Weight Sparse}

When the critical areas of our heatmap are not continuous but are separated by some non-important areas, the Weight Sparse problem occurs. This is the case where the UAV is in an area where the average $r$ of the uncovered reachable eight-neighborhood vertices of the current vertex is lower than a certain $r_{th} $ and thus, the local solutions are not good enough. Weight Sparse reduces the search efficiency because a UAV has a poor probability of covering that area efficiently. For example, it is unreasonable to search in the water for people trapped by fire, and it is harmful to people who need to be rescued on land.  An example is depicted in \Cref{fig:Problems Overview} (c) above.

\par Similarly, SVeC can also solve this type of problem. We designate UAV's next target $\eta_{nxt}$, which is the nearest vertex whose $r$ is greater than a certain threshold $r_{th}$ ( \Cref{fig:Problems Overview} (c) below). In this way, the accumulated Eff-weight can be ensured as good as possible in the case of limited endurance because there is no need to consume UAV energy in areas with no relatively important vertices.

\subsection{Weight Redundancy}

When the above three problems are solved, Weight Redundancy may affect the quality of the local solutions because a UAV often faces several uncovered reachable vertices in the eight-neighborhood having the same $r$ (\Cref{fig:Problems Overview} (d) above). These local solutions may be equivalent and redundant at the current processing step, but from the perspective of the overall MAEl-CPP task, they have apparent differences.

Thus, we employ the VwRR method to solve this problem, as illustrated in  \Cref{fig:Problems Overview} (d) below with the corresponding pseudo-code presented in \Cref{fig:Algorithm Overview} Algorithm 4. The cells' average heat-value and the corresponding vertices' average Eff-weights in the green box are significantly higher than those in the blue and yellow boxes, so when $\eta_{nxt }$ is specified as the green vertex \textbf{C} in the green box, we have a higher Eff-weight in the next few coverage steps because the potential efficiency is optimized.

	  

%% file: src/RESULTS.tex
\label{sec:RESULTS&ANALYS}

This section analyses and discusses the performance of different algorithms on solving the MAEl-CPP problem. Our experiments involve both multi-UAV and single-UAV setups. The algorithms challenged are the Zig-Zag, Na-Greedy, Heu-Greedy, and SVReC, while the evaluation metrics are the cover-rate $\sigma$, accumulated Eff-weight $R$, and the time required to accumulate a certain $R$.

\subsection{Simulation Environment}
To set up the simulation environment, we adopt the UE4 \cite{ue4} and the Airsim \cite{microsoft}. Specifically, we load and edit the environment according to our task requirements in the UE4 engine and rasterize it into cells of $40m \times 40m$. Then we obtain the heatmap utilizing the method introduced in \Cref{sec:Potential Disaster heatmap}. For the CPP part, we calculate the directed subgraph $\gamma$ in the graph structure through our algorithm and then convert it into waypoints. Waypoints are sent to the UAVs for position-based control, and then the UAV performs waypoints tracking and hovers over each cell that needs to be searched. We verify that our method can generate a heatmap based on the simulation environment, and our classifier's accuracy exceeds 91\%, affording the UAVs to perform path planning successfully.

\begin{figure}[t]
\centering
\includegraphics[width = \linewidth]{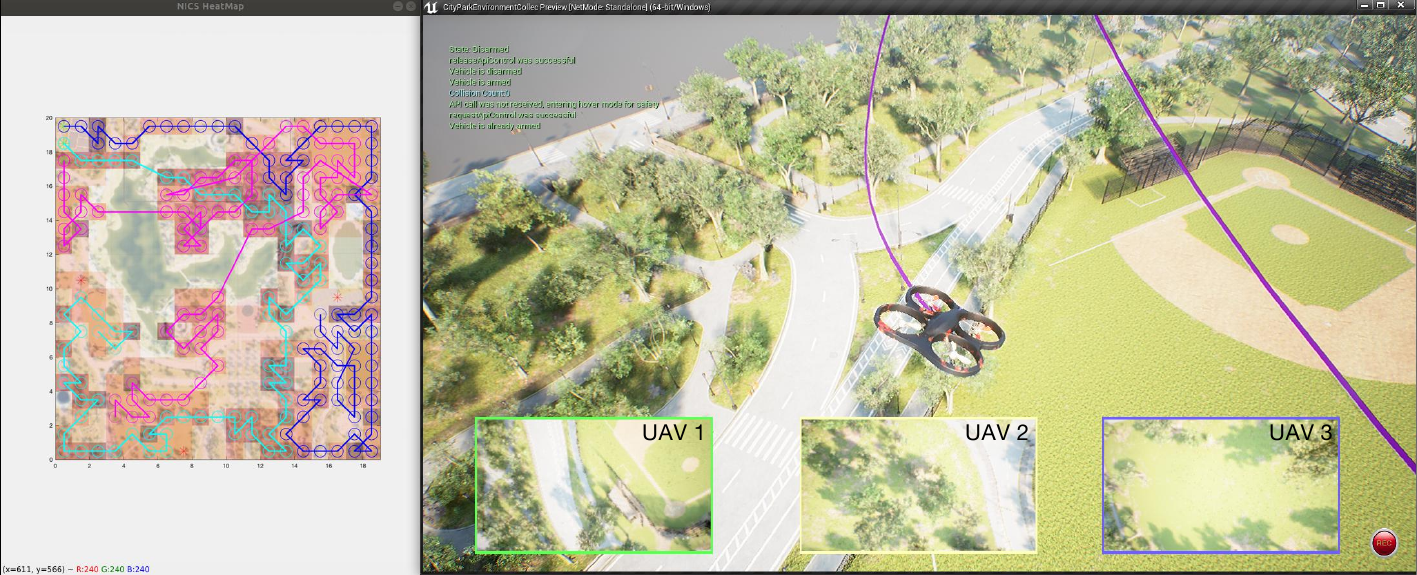}
\caption{Three UAV Simulation using SVReC in City Park}
\label{fig:airsim}
\end{figure}

The four evaluated algorithms, i.e., Zig-Zag, Na-Greedy, Heu-Greedy, and SVReC, are applied to the heatmap of the City Park \cite{unreal} simulation environment. In the simulation environment, we simultaneously display the heatmap with the coverage paths and the UAV's downward-looking cameras, along with the flight status. This simulation environment manages a greater similarity to the real environment than random heatmaps. As illustrated in \Cref{fig:airsim}, we use three UAVs and set $d_m=270$ for the simulation experiment. The SVReC algorithm covers most of the important areas in the simulation experiment.

\begin{table*}[t]
\caption{Single UAV Experimental Results}
\label{tab:single}
\begin{center}
\renewcommand{\arraystretch}{1.4}
\begin{tabular}{|c|c|c|c|c|c|c|c|c|c|c|c|c|c|}
\hline
\multicolumn{2}{|c|}{\multirow{4}{*}{UAV Number n=1}} & \multicolumn{12}{c|}{Map Size (cells)}                                                                                                                                                                  \\ \cline{3-14} 
\multicolumn{2}{|c|}{}                                & \multicolumn{4}{c|}{City Park 19x20}                              & \multicolumn{4}{c|}{50x50}                                      & \multicolumn{4}{c|}{100x100}                                      \\ \cline{3-14} 
\multicolumn{2}{|c|}{}                                & \multicolumn{2}{c|}{$d_m$ =90} & \multicolumn{2}{c|}{$d_m$ = 150} & \multicolumn{2}{c|}{$d_m$=250} & \multicolumn{2}{c|}{$d_m$=550} & \multicolumn{2}{c|}{$d_m$=1100} & \multicolumn{2}{c|}{$d_m$=2100} \\ \cline{3-14} 
\multicolumn{2}{|c|}{}                                & $\sigma$     & $\Sigma R_{i}$    & $\sigma$      & $\Sigma R_{i}$     & $\sigma$     & $\Sigma R_{i}$    & $\sigma$     & $\Sigma R_{i}$    & $\sigma$     & $\Sigma R_{i}$     & $\sigma$     & $\Sigma R_{i}$     \\ \hline
\multirow{4}{*}{Algorithm}        & Zig-Zag           & \textbf{23.7}\%       & 282.1           & \textbf{39.5}\%        & 444.4            & \textbf{10.0}\%       & 426.5           & \textbf{22.0}\%       & 976.6           & \textbf{11.0}\%       & 2034.4           & \textbf{21.0}\%       & 3616.9           \\ \cline{2-14} 
                                  & Na-Greedy         & 21.1\%       & 396.6           & 34.5\%        & 596.8            & 7.1\%        & 672.9           & 5.3\%        & 476.4           & 3.0\%        & 1200.0           & 4.0\%        & 1578.5           \\ \cline{2-14} 
                                  & Heu-Greedy        & 21.1\%       & 398.6           & 34.5\%        & 660.3            & 8.2\%        & 760.4           & 17.6\%       & 1174.4          & 8.8\%        & 2612.9           & 18.3\%       & 4182.6           \\ \cline{2-14} 
                                  & SVReC             & 19.7\%       & \textbf{436.5}           & 31.8\%        & \textbf{698.9}            & 7.2\%        & \textbf{988.2}           & 15.2\%       & \textbf{1911.0}          & 7.4\%        & \textbf{3976.5}           & 14.2\%       & \textbf{7210.6}           \\ \hline
\end{tabular}
\end{center}
\end{table*}

\begin{table*}[t]
\caption{Multi-UAV Experimental Results}
\label{tab:multi}

\begin{center}
\renewcommand{\arraystretch}{1.4}

\begin{tabular}{|c|c|c|c|c|c|c|c|c|c|c|c|c|c|}
\hline
\multicolumn{2}{|c|}{\multirow{4}{*}{UAV Number n=6}} & \multicolumn{12}{c|}{Map Size (cells)}                                                                                                                                                                    \\ \cline{3-14} 
\multicolumn{2}{|c|}{}                                & \multicolumn{4}{c|}{City Park 19x20}                               & \multicolumn{4}{c|}{50x50}                                       & \multicolumn{4}{c|}{100x100}                                      \\ \cline{3-14} 
\multicolumn{2}{|c|}{}                                & \multicolumn{2}{c|}{$d_m$ =180} & \multicolumn{2}{c|}{$d_m$ = 300} & \multicolumn{2}{c|}{$d_m$=450} & \multicolumn{2}{c|}{$d_m$=1050} & \multicolumn{2}{c|}{$d_m$=1800} & \multicolumn{2}{c|}{$d_m$=4200} \\ \cline{3-14} 
\multicolumn{2}{|c|}{}                                & $\sigma$     & $\Sigma R_{i}$     & $\sigma$      & $\Sigma R_{i}$     & $\sigma$     & $\Sigma R_{i}$    & $\sigma$     & $\Sigma R_{i}$     & $\sigma$     & $\Sigma R_{i}$     & $\sigma$     & $\Sigma R_{i}$     \\ \hline
\multirow{4}{*}{Algorithm}        & Zig-Zag           & \textbf{57.3}\%       & 652.4            & \textbf{67.7}\%        & 920.7            & \textbf{25.7}\%       & 1091.0          & \textbf{33.7}\%       & 1348.4           & \textbf{12.9}\%       & 2216.6           & 16.8\%       & 2780.9           \\ \cline{2-14} 
                                  & Na-Greedy         & 41.0\%       & 665.3            & 67.2\%        & 991.0            & 14.7\%       & 934.6           & 27.0\%       & 1691.9           & 4.6\%        & 1130.6           & 6.6\%        & 1752.3           \\ \cline{2-14} 
                                  & Heu-Greedy        & 40.1\%       & 644.5            & 59.1\%        & 892.0            & 14.1\%       & 889.0           & 24.4\%       & 1495.9           & 11.4\%       & 2532.4           & 25.2\%       & 5768.7           \\ \cline{2-14} 
                                  & SVReC             & 38.9\%       & \textbf{709.8}            & 61.7\%        & \textbf{1086.0}           & 13.9\%       & \textbf{1101.9}          & 29.0\%       & \textbf{2466.1}           & 12.8\%       & \textbf{4189.7}           & \textbf{28.0}\%       & \textbf{9158.5}           \\ \hline
\end{tabular}
\end{center}
\end{table*}

\begin{table}[t]\Huge
\caption{Time Consuming for Algorithms to accumulate a certain $R$}
\resizebox{\linewidth}{!}{
\begin{tabular}{|l|l|l|l|l|l|l|l|l|l|}
\hline
\multicolumn{2}{|l|}{n}          & \multicolumn{4}{l|}{1}         & \multicolumn{4}{l|}{6}         \\ \hline
\multicolumn{2}{|l|}{$R$}        & 200   & 400   & 600   & 800    & 200   & 400   & 600   & 800    \\ \hline
\multicolumn{2}{|l|}{Zig-Zag}    & 264.3 & 644.0 & 933.3 & 1270.0 & 240.6 & 574.2 & 844.4 & 1076.7 \\ \hline
\multicolumn{2}{|l|}{Na-Greedy}  & 157.6 & 405.6 & /     & /      & 53.9  & 85.6  & 124.2 & 170.1  \\ \hline
\multicolumn{2}{|l|}{Heu-Greedy} & 157.6 & 405.6 & 603.1 & 810.9  & 58.4  & 85.1  & 126.4 & 178    \\ \hline
\multicolumn{2}{|l|}{SVREC}      & 157.6 & 340.6 & 535.8 & 766.1  & 44.4  & 71.1  & 103.7 & 146.4  \\ \hline
\end{tabular}
}
\label{tab:speed}
\end{table}

\subsection{Eff-weight Evaluation} 
For the Eff-weight Evaluation, the experimental settings per algorithms are shown in \Cref{tab:single,tab:multi}. In this trial we test the algorithms on two cases: $n=1$ and $n=6$. In the City Park heatmap, a single UAV is set to $d_m=90$ and $d_m=150$, while for the multi-UAVs scenario, $d_m=180$ and $d_m=300$. To further evaluate our algorithm's capabilities, we randomly generate several heatmaps. Although the latter heatmaps do not have a typical regional distribution like the simulation environment, they have a larger size, simulating extreme situations. For the trials, we utilize random heatmaps of sizes 50x50 and 100x100. If the grid scale is the same as the grid map of the 
simulation environment, the 100x100 heatmap mapped to the real environment is 16 square kilometers. In these maps, a single UAV is set to $d_m=250$, $d_m=550$, $d_m=1100$ and $d_m=2100$, while when multiple UAVs are involved, then $d_m=450$, $d_m=1050$, $d_m=1800$ and $d_m=4200$. For all maps, the endurance load of each UAV in the multi-UAV case is one-sixth of the $d_m$. A UAV starts from a randomly generated start vertex in each trial, and all algorithms use the same random start vertex in each experimental round. Once the coverage is over, we record the coverage rate and the cumulative Eff-weight.

\textbf{Comparison}: As highlighted in the columns under the label City Park in \Cref{tab:single,tab:multi}, when the map size is small, Heu-Greedy and SVReC perform better than Zig-Zag and Na-Greedy. SVReC adopts a logic that focuses more on potential trends and is more efficient in searching. So it performs better than its competitor algorithms. Although in some cases, the coverage rate of SVReC is slightly lower than the other algorithms, it still achieves a higher search efficiency demonstrating its efficiency in solving the Weight Redundancy and Weight Sparse problems. Therefore it can increase the probability of survivors being rescued and reduce the loss of life and property. Considering the random large-sized heatmaps, the columns under label $50\times50$ and $100\times100$ in \Cref{tab:single,tab:multi} highlight that the Na-Greedy algorithm performs poor due to the Weight Trap and Weight Interference, significantly reducing its coverage rate. Although the Heu-Greedy method successfully solves Weight Interference and avoids stopping the coverage, it suffers from the Weight Trap when $r$ in the local optimal solution area alternately drops and invalid coverage is repeated between several vertices. Comparing both  \Cref{tab:single,tab:multi}, we conclude that as the map size increases, SVReC demonstrates a more obvious advantage in $R$ compared to its performance in the City Park map, which is significantly higher than the other three algorithms. Its coverage rate is also higher than that of Na-Greedy and Heu-Greedy. This is because as the map size increases, SVReC avoids various problems that may lead to a stop of coverage or invalid coverage. When endurances are close, the average $R$ per unit endurance of the multi-UAV SVReC is slightly lower than the single-UAV SVReC, but designing a single UAV set up with extremely long endurance is much more difficult than assigning these load endurances to multiple UAVs. Therefore, we propose the MAEl-CPP problem and the algorithms presented above.

\subsection{Searching Time Evaluation }
\label{subsec:Searching Time Evaluation}


Regarding the evaluation of the searching time, we perform a speed racing experiment in the City Park environment and record per algorithm the average time in seconds required to accumulate a certain accumulated Eff-weight $R$. Our trials consider both the single-UAV and multi-UAV cases. The UAV's speed is set to $15m/s$, while for each cell center within the trajectory, we set a hovering time of 2 seconds. The corresponding results are presented in \Cref{tab:speed}.

\textbf{Comparison}: Comparing the last three columns with the third one in \Cref{tab:speed}, we conclude that Na-Greedy, Heu-Greedy, and SVReC require significantly less time than Zig-Zag for the same $R$,  in both single-UAV and multi-UAV setups. Nevertheless, the single-UAV Na-Greedy cannot accumulate an Eff-weight more than 596 due to Weight Trap, Weight Sparse, and Weight Redundancy, as shown in the third and fourth columns of the Na-Greedy method. Under all conditions, SVReC performs better than the remaining algorithms, proving that the Eff-weight definition and SVReC's internal logic are reasonable. It is also worth noting that by comparing the columns under labels 1 and 6 for the MAEl-CPP problem, the multi-UAV case accumulates the same Eff-weight in a much shorter time than the single-UAV.

%% file: src/conclusion.tex
\label{sec:CONCLUSION}
This paper proposes a new MAEl-CPP problem with explicit energy constraints and efficiency definitions, along with a heatmap that characterizes potential disaster risks and several MAEl-CPP algorithms. After several trials followed by analysis, the proposed SVREC algorithm  is significantly better than the Na-Greedy, Heu-Greedy, and Zig-Zag algorithms when the endurance is limited. Additionally, SVREC is 4.2 times as efficient as the existing algorithm in terms of Eff-weight on large size maps.

\par However, our algorithm also has certain limitations. For example, for this type of NP-Hard problem, our algorithm's strategy cannot be guaranteed optimal. Secondly, our search and rescue algorithm  is currently based on an offline \emph{a priori} known environment and does not have online path planning capabilities if the UAV's downward-looking camera is fixed.

\par Future work shall involve deploying our algorithm on actual robots and terrain environments. Additionally, we will also focus on combining Micro-Macro control \cite{SHARON1993209} on the UAV, aiming for more practical search and rescue actions.
